\documentclass[letterpaper]{article}

\usepackage{aaai}
\usepackage{times}
\usepackage{helvet}
\usepackage{courier}

\usepackage{graphicx}
\usepackage{booktabs}
\usepackage{multirow}
\usepackage{tabularx}
\usepackage{adjustbox}
\usepackage{placeins}
\usepackage{xcolor}
\usepackage{comment}
\usepackage{tikz}
\usetikzlibrary{arrows.meta, positioning, shapes.geometric}
\title{Understanding Clinical Cognitive Dialogues Using Large Language Models}

\author{
    {\fontsize{12}{15}\selectfont\bfseries
    Vishalakshi Arumugam$^1$, Dan Schumacher$^1$, Veronica Rammouz$^1$, $^1$Erfan Nourbakhsh$^1$,}\\
    {\fontsize{12}{15}\selectfont\bfseries
     Enrique Gonzalez Guerrero$^2$, Jeremy Davis$^1$, and Anthony Rios$^1$}\\
    $^1$University of Texas at San Antonio, USA\\
    $^2$Tec de Monterrey, MX\\
    \texttt{\{vishalakshi.arumugam, anthony.rios\}@utsa.edu}
}

\begin{document}

\maketitle

\begin{abstract}
In-person cognitive assessment is both a test and an interaction. Clinicians explain tasks, repair misunderstandings, and adapt to patient responses, while patients may hesitate, seek clarification, or disengage. Yet clinical dialogue resources rarely label the interaction structure needed to study these behaviors at scale. We present an de-identified corpus of 33 cognitive assessment conversations with 8,250 utterances annotated for three speaker roles and 56 dialogue acts. We use this corpus to benchmark large language models on fine-grained dialogue-act classification and next-patient-utterance generation. We also test whether out-of-domain instruction data and explanation-augmented training transfer to this clinical setting. Instruction tuning produces the strongest patient-utterance reference matching and improves classification accuracy. Reasoning-aware fine-tuning produces the strongest classification results among the LLaMA-3.1-8B variants. However, even the best models struggle to separate closely related dialogue acts, showing that broad conversational intent is easier to recognize than fine-grained communicative function. The corpus and benchmark make interaction structure measurable in cognitive assessments and support follow-up work on conversational markers, clinician education, and carefully validated simulated patients. This work does not make diagnostic claims. Instead, it provides the data and evaluation framework needed to study these applications.
\end{abstract}
\section{Introduction}

In-person cognitive assessments are used to evaluate memory, reasoning, language, and executive function, especially among older adults who may have cognitive impairment. These assessments are not simply collections of questions and scored answers. They unfold through interaction. Clinicians explain tasks, reformulate questions, check understanding, and respond to confusion, while patients may hesitate, request clarification, revise an answer, or rely on an accompanying family member. Variation in how a cognitive test is administered can affect patients' understanding and responses to tests~\cite{jones2019variation}. These interactions are part of the assessment process, but they are difficult to study across long, multi-party conversations.

Most computational work on cognitive impairment has focused on acoustic or linguistic markers in patient speech and on predicting cognitive or diagnostic categories~\cite{luzAlzheimersDementiaRecognition2020}. A recent review found that NLP research on dementia remains centered on detection and called for broader datasets, tasks, and clinically grounded applications~\cite{peled-cohen-reichart-2025-systematic}. Final predictions alone do not capture how an assessment was conducted, how clinicians and patients responded to each other, or where misunderstanding and conversational repair occurred. Studying these interaction patterns could support research on assessment administration, patient participation, and associations between conversation structure and cognitive outcomes. It could also inform training materials for clinicians and neuropsychologists. There is a need is an annotated resource and reliable methods for measuring these interactions.

Dialogue acts provide a structured representation of conversational function. They describe whether an utterance asks a question, provides information, acknowledges a response, or requests clarification~\cite{searle-1969-speech-acts,stolcke-etal-2000-dialogue}. Later frameworks extended dialogue-act annotation to represent communicative intent, turn management, and discourse structure~\cite{core-allen-1997-damsl,popescu-belis-2005-dialogue,bunt2012iso}. Work on general conversations and meetings has shown that context, speaker roles, and turn structure are important for recognizing these functions~\cite{shribergICSIMeetingRecorder2004,zelasko-etal-2021-helps,he-etal-2021-speaker-turn,qamar-etal-2023-speaking}.

Clinical dialogue research has applied related methods to distress detection, dementia recognition, counseling, and conversational behavior~\cite{gratchDistressAnalysisInterview2014,malhotraSpeakerTimeawareJoint2021,hoxhaDREAMClassificationScheme2016}. However, existing resources focus on general interviews, counseling conversations, or short structured tasks. Cognitive assessments present a different setting. They contain repeated procedures, closely related question types, short patient responses, changing clinician strategies, and subtle differences in communicative intent. Caregivers or family members may also participate. Existing public resources rarely combine fine-grained dialogue-act labels, explicit speaker roles, surrounding context, and multi-party cognitive assessment conversations.

Models of interaction structure could eventually support simulated-patient systems for education. Virtual patients can provide controlled scenarios for practicing clinical reasoning and communication, but they should complement rather than replace contact with real patients and must be carefully validated~\cite{kononowicz2019virtual}. Before such systems can be considered, we must determine whether models can recognize communicative intent and generate contextually appropriate responses. We therefore study patient utterance generation as a controlled benchmark of contextual response modeling, not as evidence that a model can replace a patient or accurately simulate cognitive impairment.

We introduce an de-identified corpus of 33 in-person cognitive assessment conversations containing 28,592 utterances. Of these, 8,250 are annotated with three speaker roles and 56 fine-grained dialogue acts. Three annotators independently labeled the data, with difficult cases adjudicated by a board-certified clinical neuropsychologist. We use this resource to study two related tasks. Dialogue-act classification tests whether models can distinguish closely related communicative functions. Patient utterance generation tests whether they can produce a response that fits the preceding conversation and available patient metadata.

We compare prompting, instruction tuning, and reasoning-aware fine-tuning. For training-based adaptation, we construct a 41,836-example corpus covering dialogue understanding, question answering, summarization, behavioral dialogue, emotion reasoning, and instruction following. We also create a reasoning-enhanced version with teacher-generated explanations. The results show that adaptation effects depend on the task and metric. Instruction tuning produces the strongest patient-utterance reference matching and improves classification accuracy. Reasoning-aware fine-tuning provides the clearest classification gains for LLaMA-3.1-8B. However, all models remain much better at recognizing broad categories such as statements, questions, and feedback than at separating fine-grained functions within those categories.

The immediate contribution of this work is research infrastructure rather than a clinical system. The corpus supports future study of dialogue-act sequences, conversational repair, speaker participation, and their relationships with assessment outcomes. To facilitate follow-up work, we will release the annotation guidelines, label definitions, patient-level task splits, prompts, training and evaluation code, and model outputs. Access to de-identified transcripts will follow Institutional Review Board requirements and applicable data-use agreements. 

Our contributions are:
\textbf{1)} We introduce a de-identified corpus of 33 cognitive assessment conversations with 8,250 utterances annotated for three speaker roles and 56 fine-grained dialogue acts.
\textbf{2)} We establish benchmarks for fine-grained dialogue-act classification and context-based patient utterance generation in cognitive assessment conversations.
\textbf{3)} We construct a 41,836-example multi-task instruction corpus and a reasoning-enhanced version to study transfer from out-of-domain dialogue supervision.
\textbf{4)} We show that adaptation methods have task-dependent effects and identify a central limitation of current models: broad conversational intent transfers more reliably than fine-grained communicative function.
\textbf{5)} We provide a release plan for the annotation resources, task splits, prompts, code, and model outputs to support reproducible follow-up research.

\section{Related Work}

\noindent \textbf{Clinical Dialogue Understanding.}
Dialogue acts describe the purpose of an utterance, such as asking a question, giving information, acknowledging a response, or requesting clarification~\cite{searle-1969-speech-acts,farzanaModelingDialogueConversational2020}. General-domain datasets such as Switchboard, ICSI, HCRC Map Task, and AMI established common annotation schemes and benchmarks for dialogue-act classification~\cite{jurafsky1997ws,janinICSIMeetingCorpus2003,shribergICSIMeetingRecorder2004,carlettaReliabilityDialogueStructure1997,carlettaAMIMeetingCorpus2006}. Clinical resources later extended dialogue analysis to dementia assessment, clinical interviews, and mental health counseling~\cite{luzAlzheimersDementiaRecognition2020,pope2011finding,gratchDistressAnalysisInterview2014,malhotraSpeakerTimeawareJoint2021}.

Several frameworks capture the structure of clinical communication. Bunt et al. (2012) represents dialogue acts across semantic dimensions~\cite{bunt2012iso}, DREAM adapts DAMSL to clinical research interactions~\cite{hoxhaDREAMClassificationScheme2016}, and RIAS models task, relational, and affective functions in physician--patient conversations~\cite{roter2002roter}. We use the fine-grained MRDA taxonomy as the basis for our annotation scheme~\cite{dhillonMeetingRecorderProject2004}. Dialogue-act methods have progressed from statistical models using lexical, prosodic, and contextual features~\cite{stolcke-etal-2000-dialogue} to neural models that capture longer conversational context~\cite{nasreenRareClassDialogueAct2021}. However, current language models still struggle to distinguish closely related dialogue acts~\cite{qamarLLMsUnderstandDialogues2025a}. Clinical studies have also used smaller dialogue-act sets for dialogue management, cognitive screening, and dementia detection~\cite{guptaBuildingVirtualAssistant2018a,farzanaModelingDialogueConversational2020,liAlzheimersDementiaDetection2022}. Our work instead studies 56 fine-grained dialogue acts in structured cognitive assessments and evaluates classification and patient utterance generation.

\vspace{.2mm}
\noindent \textbf{Instruction Tuning.}
Instruction tuning adapts pretrained models using collections of instruction--response pairs and can improve generalization across tasks~\cite{wei2021finetuned}. Related work combines supervised tuning with human feedback to improve instruction following~\cite{ouyang2022training}. Self-Instruct reduces the need for manually written examples by generating new instruction data automatically~\cite{wang-etal-2023-self-instruct}, while Alpaca showed that teacher-generated instruction data can support effective adaptation of open models~\cite{taori2023alpaca}. These methods have been studied mainly on general tasks such as question answering and summarization~\cite{wei2021finetuned,wang-etal-2023-self-instruct,taori2023alpaca}. We test whether dialogue-focused instruction data transfers to structured clinical conversations and whether it improves both dialogue-act classification and patient utterance generation.

\vspace{.2mm}
\noindent \textbf{Reasoning-Aware~Fine-Tuning.}~Chain-of-Thought prompting asks a model to produce intermediate reasoning before its answer~\cite{wei2022chain}. STaR extends this idea by training models on generated reasoning traces~\cite{zelikman2022star}, while Orca uses explanations from a stronger teacher model as supervision~\cite{mukherjee2023orca}. Most prior work evaluates reasoning supervision on mathematics, question answering, or coding. Its value for dialogue understanding is less clear, especially when intent depends on speaker roles and prior turns. We compare prompting, instruction tuning, and reasoning-aware fine-tuning under the same clinical setting to determine whether their effects differ across classification and generation.

\section{Methodology}

\begin{figure*}[t]
    \centering
    \includegraphics[width=\textwidth]{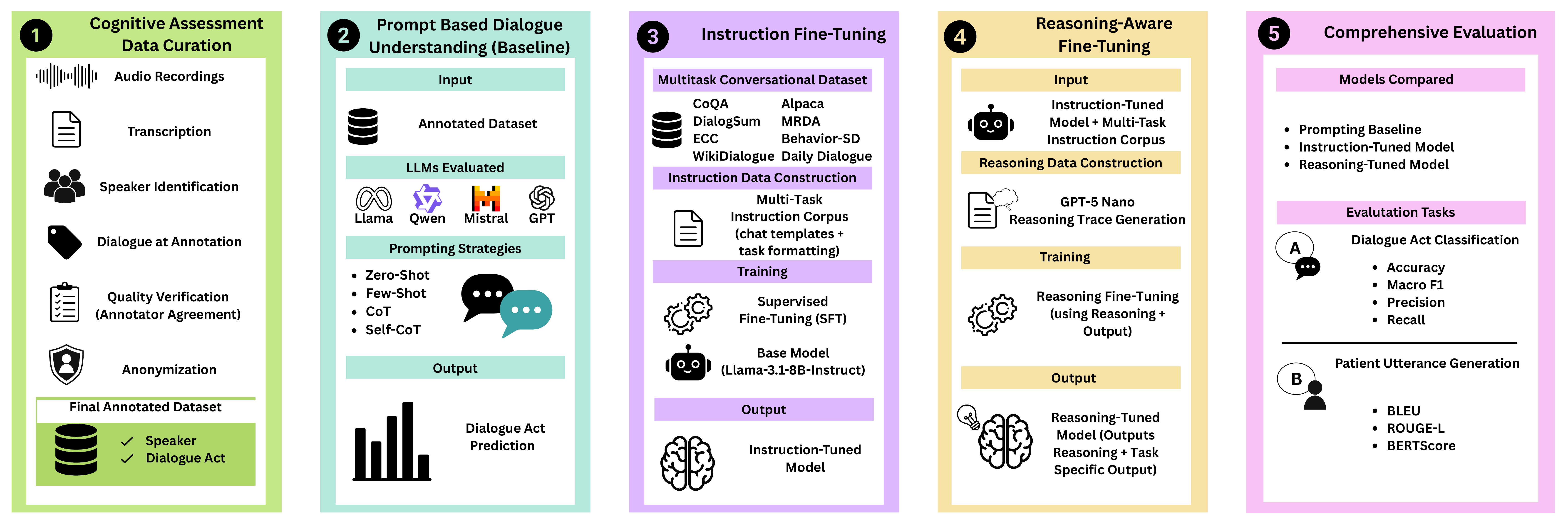}
    \caption{Overview of the data construction, model adaptation, and evaluation tasks.}
    \label{fig:methodology}
\end{figure*}

Figure~\ref{fig:methodology} summarizes how we construct an annotated corpus and train models.

\subsection{Clinical Cognitive Assessment Corpus}

\paragraph{Data Collection and Transcription.}
The corpus contains 33 in-person cognitive assessments collected through a large university hospital system. Each conversation includes a clinician and patient and may also include a caregiver or family member. Abridge's digital scribe system produced the initial transcripts. Four Ph.D. students manually corrected transcription and speaker errors with guidance from a board-certified clinical neuropsychologist (ABPP). All transcripts were de-identified before annotation. The full corpus contains 28,592 utterances. From these conversations, 8,250 utterances were selected for annotation by selecting consecutive sections focused on cognitive testing while excluding general conversation that could disclose PII. The Supplementary Material provides further corpus characterization, including clinician participation, accompanying persons, cognitive-status groups, assessment procedures, conversation lengths, recruitment criteria, and repeated clinicians or assessment templates, as well as additional details on de-identification and controlled data access. Table~\ref{tab:corpus-stats} reports the corpus statistics.

\begin{table}[t]
\centering
\setlength{\tabcolsep}{6pt}
\renewcommand{\arraystretch}{1.15}
\begin{adjustbox}{width=.6\linewidth}
\begin{tabular}{lr}
\toprule
\textbf{Statistic} & \textbf{Value} \\
\midrule
Number of Patients & 33 \\
Total Utterances & 28{,}592 \\
Annotated Utterances & 8{,}250 \\
Average Utterance Length (words) & 5.12 \\
Standard Deviation of Length & 5.19 \\
Speaker Roles & 3 \\
Dialogue-Act Labels & 56 \\
\bottomrule
\end{tabular}
\end{adjustbox}
\caption{Statistics of the cognitive assessment corpus.}
\label{tab:corpus-stats}
\end{table}

\vspace{.2mm}
\noindent \textbf{Annotation.}
Each utterance received two labels: a speaker role and a dialogue act. The speaker roles are \textsc{Clinician}, \textsc{Patient}, and \textsc{Accompanying Person}. The dialogue-act schema contains 56 labels based on the Meeting Recorder Dialogue Act framework~\cite{dhillonMeetingRecorderProject2004}. This taxonomy is important because several labels capture communication behaviors with direct clinical
relevance. \textsc{Understanding Check} and clarification or repair
capture efforts to establish shared understanding; memory-clinic
clinicians often fail to check understanding, while patient-initiated
clarification has been associated with later treatment
adherence~\cite{visser2019clinician,mccabe2013shared}. \textsc{Open-Ended Question},
\textsc{Elaboration}, and fine-grained question labels capture
information elicitation and response structure, which are associated
with patient disclosure and distinguish conversational profiles in
memory-clinic consultations~\cite{roter1987physicians,jones2016conversational}. \textsc{Third-Party Talk} and the
\textsc{Accompanying Person} role capture triadic communication, where
companions may assume a larger role when patient understanding is
uncertain~\cite{windeatt2026doing}.
Table~\ref{tab:example-anno} provides examples.
Three Ph.D. student annotators independently labeled all 8,250 utterances. A custom Streamlit interface (See Supplementary Material) displayed the target utterance together with the preceding and following turns. This allowed annotators to use the surrounding context when assigning labels.

\begin{table}[t]
\centering
\setlength{\tabcolsep}{5pt}
\renewcommand{\arraystretch}{1.15}
\begin{adjustbox}{width=.9\linewidth}
\begin{tabular}{lll}
\toprule
\textbf{Utterance} & \textbf{Speaker} & \textbf{Dialogue Act} \\
\midrule
Have you seen audiology? & Clinician & Wh-Question \\
No & Patient & Negative Answer \\
Yeah & Accompanying Person & Acknowledgement \\
\bottomrule
\end{tabular}
\end{adjustbox}
\caption{Examples with speaker-role and dialogue-act labels.}
\label{tab:example-anno}
\end{table}

We assessed inter-annotator agreement separately at the corpus and label levels. For dialogue acts, naïve agreement was \(71.42\%\) and Fleiss' \(\kappa=.457\); for speaker roles, naïve agreement was \(92.02\%\) and Fleiss' \(\kappa=.713\). The Supplementary Material reports support and mean pairwise Cohen's \(\kappa\) for every label. Speaker-role agreement was substantial for \textsc{Clinician} (\(\kappa=.794\)) and \textsc{Patient} (\(\kappa=.756\)), and moderate for \textsc{Accompanying Person} (\(n=562,\ \kappa=.552\)). Dialogue-act reliability varied across the taxonomy: \textsc{Negative Answer} (\(n=441,\ \kappa=.858\)), \textsc{Y/N Question} (\(n=946,\ \kappa=.697\)), and \textsc{Backchannel} (\(n=909,\ \kappa=.602\)) showed comparatively strong agreement, whereas \textsc{Defending/Explanation} (\(n=479,\ \kappa=.203\)) was more difficult to distinguish. Across the 56 dialogue acts \(\kappa\) was as high as \(.858\).

Three annotators independently assigned speaker roles and dialogue acts using the MRDA-based definitions, annotation guidelines, and surrounding dialogue context. When two annotators selected the same label, that majority label was retained; cases without a majority or requiring clinical interpretation were reviewed by the board-certified clinical neuropsychologist. We refer to the resulting annotations as \emph{adjudicated consensus labels}, which operationalize the annotation scheme rather than establish a uniquely correct interpretation. Representative cases included \textsc{Statement} versus \textsc{Elaboration} or \textsc{Defending/Explanation}, \textsc{Acknowledgement} versus \textsc{Backchannel} or \textsc{Accept}, and \textsc{Y/N Question} versus \textsc{Understanding Check}. These also motivate our family-level error analysis. Accordingly, we report label-wise reliability, emphasize Macro-F1 under class imbalance, and interpret performance on rare or low-agreement labels as agreement with the operational taxonomy rather than error against unambiguous ground truth. This treatment is consistent with the documented difficulty of fine-grained dialogue-act annotation~\cite{xu2023medical,cai2025search}.

\subsection{Model Training}

\noindent \textbf{Prompting Baselines.}
For dialogue-act classification, each model receives the target utterance, its dialogue context, the complete set of labels, and the annotation guidelines. The model must return exactly one of the 56 labels. We use the same instructions and output format for all models. We evaluated zero-shot, few-shot, and CoT prompting. 

\vspace{.2mm}
\noindent \textbf{Instruction Tuning.}
We construct a multi-task instruction corpus from eight public datasets: Alpaca, DailyDialog, ECC, Behavior-SD, WikiDialogue, CoQA, MRDA, and DialogSum. The final corpus contains 41,836 instruction--response pairs. Table~\ref{tab:instruction-corpus} reports its composition.
The datasets were selected to cover skills needed for dialogue modeling, including instruction following, dialogue-act recognition, question answering, summarization, emotion reasoning, and conversational response generation. These datasets do not reproduce the clinical data distribution. Instead, they allow us to test whether general knowledge of dialogue transfers to cognitive assessment conversations.

We convert each example into the same format with three parts: an instruction, an input, and a target response. We preserve the original labels and reference responses from each dataset. No examples from the clinical evaluation corpus are included in the instruction-tuning data. Any improvement therefore reflects transfer from out-of-domain supervision rather than training on the evaluation conversations.

\begin{table}[t]
\centering
\setlength{\tabcolsep}{4pt}
\renewcommand{\arraystretch}{1.1}
\resizebox{.7\linewidth}{!}{%
\begin{tabular}{llr}
\toprule
\textbf{Dataset} & \textbf{Task} & \textbf{Records} \\
\midrule
Alpaca       & Instruction Following       & 25{,}000 \\
DailyDialog  & General Conversation        & 4{,}500  \\
ECC          & Emotion and Cause Reasoning & 3{,}836  \\
Behavior-SD  & Behavioral Dialogue         & 2{,}500  \\
WikiDialogue & Dialogue Act                & 2{,}500  \\
CoQA         & Question Answering           & 2{,}000  \\
MRDA         & Meeting Dialogue Acts        & 1{,}000  \\
DialogSum    & Dialogue Summarization       & 500      \\
\midrule
\textbf{Total} & \textbf{Instruction Corpus} & \textbf{41{,}836} \\
\bottomrule
\end{tabular}}
\caption{Composition of the multi-task instruction corpus.}
\label{tab:instruction-corpus}
\end{table}

\vspace{.2mm}
\noindent \textbf{Reasoning-Aware Fine-Tuning.}
We create a second version of the instruction corpus with teacher-generated explanations. For each example, GPT-5-nano receives the instruction, input, and correct target response. It produces a short explanation of the evidence supporting the response. The training target contains the explanation followed by the original response.
We evaluate two reasoning-aware models. The first is trained from the base LLaMA-3.1-8B model. The second begins from the instruction-tuned LLaMA-3.1-8B checkpoint. This design separates the effect of reasoning supervision from the effect of instruction tuning.
The explanations are used only as training supervision. We do not treat them as faithful descriptions of a model's internal reasoning. We also do not generate reasoning traces from the clinical evaluation corpus.

\subsection{Evaluation Tasks}

\noindent \textbf{Dialogue-Act Classification.}
Dialogue-act classification is a 56-class prediction task. For each target utterance, the model receives the surrounding dialogue and predicts one dialogue-act label. This task measures whether the model can distinguish fine-grained communicative functions.

\vspace{.2mm}
\noindent \textbf{Patient Utterance Generation.}
Patient utterance generation tests whether a model can generate the next patient utterance from the preceding dialogue. Let a conversation be
$
C=(x_1,\ldots,x_T),
$
where each turn $x_t=(s_t,w_t)$ contains a speaker role $s_t$ and utterance $w_t$. For each patient turn, the input contains the preceding $h$ turns, where $h\in\{1,\ldots,5\}$, and conversation-level metadata $m$. The metadata include cognitive status and whether an accompanying person is present. The task is
$
p(w_t \mid x_{t-h:t-1},m).
$
The input presents the metadata followed by the dialogue turns in chronological order using the format \textit{Speaker: Utterance}. The model is instructed to return only the next patient utterance.

This task measures how well the generated text matches the observed patient response given the same context and metadata. It does not test whether the model can replace a patient or produce clinically valid responses for direct use.

\section{Results}

\paragraph{Experimental Protocol.}
All 33 clinical conversations (8,250 annotated utterances) were reserved exclusively for evaluation; model training used only the 41,836-example out-of-domain instruction corpus, and no clinical utterance was included in model adaptation. For dialogue-act classification, all models used the same task instructions, label definitions, dialogue-context construction, fixed few-shot demonstrations drawn outside the clinical corpus, and exact-label output parsing, with nonconforming responses counted as invalid. Zero-shot, few-shot, and Chain-of-Thought (CoT); patient-utterance generation used direct next-utterance prompting without CoT. Because utterances are clustered within conversations, uncertainty was estimated using conversation-level bootstrap resampling. Complete prompts, demonstration counts and provenance, context-window definitions, parsing rules, decoding and training hyperparameters, and bootstrap procedures are provided in the Supplementary Material.

\vspace{.2mm}
\noindent \textbf{Dialogue-Act Classification.}
Table~\ref{tab:overall_results} reports dialogue-act classification results for all models and prompting strategies. Because the dataset has a highly imbalanced label distribution, we treat Macro-F1 as the main metric. Accuracy and Weighted-F1 provide additional measures that give more weight to common dialogue acts.
The strongest result depends on the metric. Mistral3-24B with few-shot prompting obtains the highest Macro-F1 at .207. Qwen3-30B with instruction tuning obtains the highest Accuracy at .480, while the original Qwen3-30B with CoT prompting obtains the highest Weighted-F1 at .414. These differences show that overall accuracy and balanced performance do not always improve together.

\begin{table}[t]
\centering
\resizebox{\linewidth}{!}{%
\begin{tabular}{llccc}
\toprule
\textbf{Model} & \textbf{Prompt} & \textbf{Acc.} & \textbf{mF1} & \textbf{wF1} \\
\midrule

\multicolumn{5}{l}{\textbf{Dummy Baselines}}\\
\midrule
Majority Classifier
& --- & .136 & .004 & .033\\

Random Classifier (Stratified)
& --- & .081 & .019 & .081\\

Random Classifier (Uniform)
& --- & .018 & .010 & .027\\
\midrule
\multicolumn{5}{l}{\textbf{Baseline Model}}\\
\midrule
LLaMA3.1-8B
& Zero-shot & .234 & .117 & .239\\
& Few-shot  & .248 & .099 & .188\\
& CoT       & \textbf{.292} & \textbf{.135} & \textbf{.295}\\

\addlinespace

Gemma3-12B
& Zero-shot & .382 & .151 & .326\\
& Few-shot  & \textbf{.412} & \textbf{.179} & \textbf{.349}\\
& CoT       & .411 & .165 & .342\\

\addlinespace

Qwen3-30B
& Zero-shot & .420 & .183 & .392\\
& Few-shot  & \textbf{.439} & \textbf{.194} & .408\\
& CoT       & .439 & .193 & \textbf{.414}\\

\addlinespace

Mistral3-24B
& Zero-shot & .417 & .194 & .382\\
& Few-shot  & \textbf{.450} & \textbf{.207} & \textbf{.393}\\
& CoT       & .409 & .185 & .344\\

\midrule

\multicolumn{5}{l}{\textbf{Instruction-Tuned Model}} \\
\midrule
LLaMA3.1-8B + Instruction FT
& Zero-shot & .316 & .096 & .222\\
& Few-shot  & \textbf{.337} & \textbf{.102} & \textbf{.259}\\
& CoT       & .330 & .083 & .250\\

\addlinespace

Qwen3-30B + Instruction FT
& Zero-shot & \textbf{.480} & .172 & \textbf{.407}\\
& Few-shot  & .462 & .169 & .377\\
& CoT       & \textbf{.480} & \textbf{.175} & .405\\
\addlinespace

\midrule

\multicolumn{5}{l}{\textbf{Reasoning-Aware Fine-Tuned Models}} \\
\midrule
LLaMA3.1-8B + Reasoning (Base)
& Zero-shot & \textbf{.423} & \textbf{.151} & \textbf{.357}\\
& Few-shot  & .414 & .142 & .337\\
& CoT       & .407 & .142 & .341\\

\addlinespace

LLaMA3.1-8B + Reasoning (Instruction)
& Zero-shot & .396 & .130 & .328\\
& Few-shot  & \textbf{.402} & \textbf{.158} & \textbf{.339}\\
& CoT       & .380 & .126 & .323\\

\bottomrule
\end{tabular}}
\caption{Performance comparison of prompting strategies, instruction tuning, and reasoning-aware fine-tuning for dialogue act classification.}
\label{tab:overall_results}
\end{table}

\begin{table}[t]
\centering
\resizebox{.65\linewidth}{!}{%
\begin{tabular}{lccc}
\toprule
\textbf{Model} & \textbf{Acc.} & \textbf{mF1} & \textbf{wF1} \\
\midrule
Dummy (Majority) & .139 & .004 & .034 \\
Dummy (Uniform)   & .018 & .010 & .026 \\ \midrule
Linear SVM       & .546 & .157 & .511 \\
RoBERTa-Large    & \textbf{.681} & \textbf{.266} & \textbf{.668} \\
BioMedBERT       & .654 & .164 & .612 \\
LLaMA3.1-8B + Reasoning (Base) & .418 & .113 & .342 \\

\bottomrule
\end{tabular}}
\caption{Performance comparison of prompting vs. supervised models.}
\label{tab:supervised_baselines}
\end{table}
\subsubsection{Effect of Instruction Tuning}

Instruction tuning improves overall task fit, but its effect differs across metrics. For LLaMA3.1-8B, the best Accuracy increases from .292 to .337. However, the best Macro-F1 decreases from .135 to .102. A similar pattern appears for Qwen3-30B. Instruction tuning increases its best Accuracy from .439 to .480, while its best Macro-F1 decreases from .194 to .175.

These results suggest that instruction tuning helps the models learn the task format and recognize common dialogue acts. However, it does not improve balanced performance across all 56 labels. In particular, gains in Accuracy should not be treated as evidence of stronger performance on rare dialogue acts.

\vspace{.2mm} \noindent \textbf{Comparison to Supervised Classification.} We evaluate prompting to supervised baselines in Table~\ref{tab:supervised_baselines}. In this setting, we split the entire dataset into training, validation, and test sets of size 5,000, 750, and 2,500 examples, respectively. Splits are based on patients, not individual utterances, to avoid training data leakage. Overall, we find that, even with substantial instruction tuning, reasoning training, and few-shot examples, LLMs are unable to match the performance of simple supervised baselines. This further suggests the need for future research on improved clinical dialogue understanding in LLMs.

\begin{table*}[t]
\centering
\resizebox{.75\textwidth}{!}{%
\begin{tabular}{lccccc}
\toprule
\textbf{Model} & \textbf{BLEU} & \textbf{ROUGE-1} & \textbf{ROUGE-2} & \textbf{ROUGE-L} & \textbf{BERTScore-F1} \\
\midrule
Gemma-3-12B & .021 & .140 & .033 & .139 & \textbf{.906} \\
Qwen3-30B & .025 & .168 & \textbf{.038} & .161 & .898 \\
Qwen3-30B (Instruction) & \underline{.041} & \textbf{.184} & \underline{.037} & \textbf{.181} & .901 \\
Mistral-3-24B & .018 & .157 & .023 & .153 & .874 \\
LLaMA-3.1-8B & .023 & .139 & .026 & .136 & .900 \\ \midrule
LLaMA-3.1-8B (Instruction) & \textbf{.043} & \underline{.178} & .036 & \underline{.175} & \textbf{.906} \\
LLaMA-3.1-8B (Reasoning) & .037 & .171 & .033 & .169 & .904 \\
LLaMA-3.1-8B (Reasoning + Instruction) & .035 & .176 & .030 & .174 & .905 \\
\bottomrule
\end{tabular}}
\caption{Patient utterance generation performance ($k=5$). Best results are shown in bold and second-best results are underlined.}
\label{tab:utterance_generation_results}
\end{table*}

\vspace{.2mm} \noindent \textbf{Effect of Reasoning-Aware Fine-Tuning.}
Reasoning-aware fine-tuning produces the strongest results among the LLaMA3.1-8B variants. The Reasoning (Base) model reaches .423 Accuracy and .151 Macro-F1 with zero-shot prompting. Compared with the strongest original LLaMA3.1-8B result, this represents a 13.1 percentage-point increase in Accuracy and a 1.6-point increase in Macro-F1.
The Reasoning (Instruction) model achieves the highest LLaMA Macro-F1, .158, with few-shot prompting. However, it does not consistently outperform Reasoning (Base). The results therefore support the reasoning-aware training procedure as an effective form of adaptation for LLaMA3.1-8B, but they do not show that starting from the instruction-tuned checkpoint is always better.
 The reasoning-aware LLaMA models also remain below the best larger models in Macro-F1. Mistral3-24B reaches .207, and Qwen3-30B reaches .194. Model scale and the quality of the original instruction tuning therefore remain important.

\subsubsection{Effect of Prompting}

The best prompting method varies by model. Few-shot prompting produces the highest Macro-F1 for Gemma3-12B, Qwen3-30B, Mistral3-24B, and the Reasoning (Instruction) model. CoT prompting gives the strongest original LLaMA3.1-8B result, but it reduces performance for several other models. The reasoning-aware models are less sensitive to the prompt choice than the original LLaMA3.1-8B model. For Reasoning (Base), Macro-F1 ranges from .142 to .151 across the three prompting methods. This suggests that training-based adaptation provides more stable performance than prompt changes alone. Overall, few-shot examples are often useful, whereas CoT does not consistently benefit dialogue-act classification.

Table~\ref{tab:utterance_generation_results} reports patient utterance generation results using five preceding turns. 
Instruction-tuned LLaMA3.1-8B obtains the highest BLEU score at .043 and ties for the highest BERTScore-F1 at .906. Instruction-tuned Qwen3-30B obtains the highest ROUGE-1 and ROUGE-L scores at .184 and .181. It also achieves the second-highest BLEU score of .041.
Instruction tuning improves the two models for which matched comparisons are available. For LLaMA3.1-8B, BLEU increases from .023 to .043, while ROUGE-L increases from .136 to .175. For Qwen3-30B, BLEU increases from .025 to .041, while ROUGE-L increases from .161 to .181.
Both reasoning-aware LLaMA variants also improve over the original LLaMA3.1-8B model. However, neither exceeds instruction tuning. These results indicate that the multi-task instruction corpus provides the clearest benefit for matching the observed patient responses.
BERTScore varies less than the lexical metrics. Most models obtain scores between .898 and .906. This suggests that several systems produce responses with broadly similar meanings, even when their wording differs from the recorded response.

\vspace{.2mm}
\noindent \textbf{Performance Across Cognitive-Status Groups.}
We also compare generation performance across cognitive-status groups. This analysis is descriptive because the groups contain different numbers of patient turns (See results in the Supplementary Material).
Instruction tuning improves lexical overlap for the largest groups, including Mild Cognitive Impairment and Normal cognition. The reasoning-aware models show their strongest relative performance on Dementia conversations, achieving higher ROUGE-L and BERTScore-F1 within that group.
BERTScore remains near .89--.91 across most conditions, while BLEU and ROUGE vary more across groups. This pattern is consistent with the overall results: generated responses often preserve broad meaning, but they do not always reproduce the wording of the recorded patient response. 

\begin{figure*}[t]
    \centering
    \includegraphics[width=.333\textwidth]
        {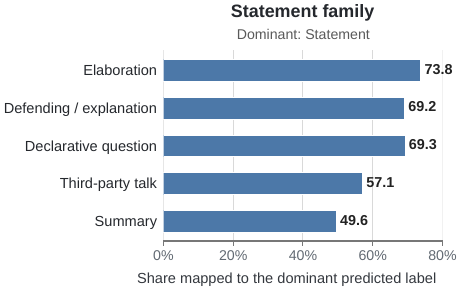}%
    \hspace{-2mm}%
    \includegraphics[width=.333\textwidth]
        {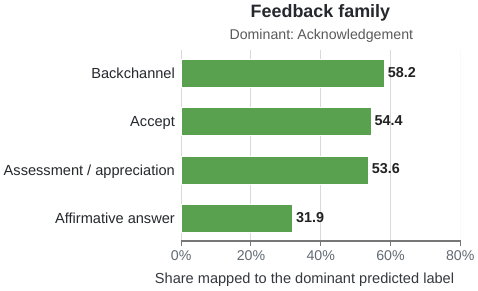}%
    \hspace{-2mm}%
    \includegraphics[width=.333\textwidth]
        {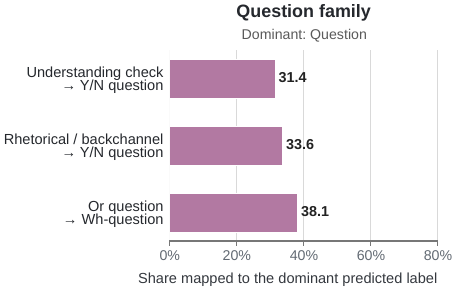}

    \vspace{-2mm}
    \caption{Dominant prediction patterns within each dialogue-act family.
    Each bar shows the percentage of examples mapped to the family's
    dominant predicted label.}
    \label{fig:error-taxonomy-families}
    \vspace{-3mm}
\end{figure*}

\vspace{.2mm} \noindent \textbf{Error Analysis.}
Aggregate metrics do not show which dialogue acts remain difficult. We therefore examine the errors shared across the 24 experimental settings, covering eight model variants and three prompting strategies. This pooled analysis identifies common classification problems across the evaluated systems. It is not designed to isolate the effect of one prompting or training method.

\vspace{.2mm} \noindent \textbf{Systematic Confusion Analysis.}
The errors follow clear patterns. Models often identify the broad purpose of an utterance but fail to select the more specific label within that dialogue-act family. Figure~\ref{fig:error-taxonomy-families} summarizes the most common confusions.
The largest errors occur within three families. In the statement family, \textit{Elaboration}, \textit{Defending/Explanation}, and \textit{Summary} are often predicted as the general \textit{Statement} label. For example, 73.8\% of errors for \textit{Elaboration} map to \textit{Statement}, as do 69.2\% of errors for \textit{Defending/Explanation}.

A similar pattern appears for feedback acts. \textit{Backchannel}, \textit{Accept}, and \textit{Assessment/Appreciation} are often predicted as \textit{Acknowledgement}. Within the question family, models confuse specific forms with broader question labels. For example, \textit{Understanding Check} is often predicted as \textit{Y/N Question}, while \textit{Or Question} is often predicted as \textit{Wh-Question}.

These errors are more often within a related dialogue-act family than across unrelated families. The models therefore capture coarse conversational intent better than fine-grained communicative function. Distinguishing these labels may require longer context, clearer label definitions, or more direct supervision for the boundary between related acts.

\begin{figure}[t]
\centering
\includegraphics[width=.7\linewidth]{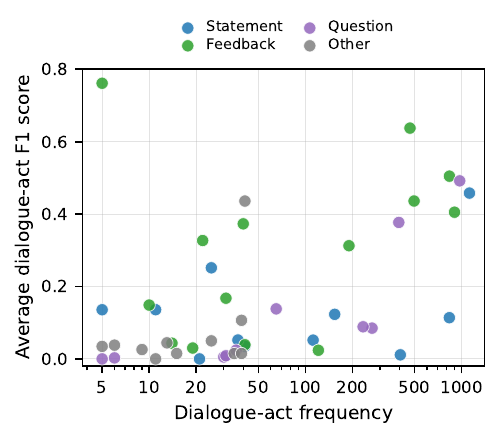}
\caption{Dialogue-act frequency versus average F1, grouped by dialogue-act family.}
\label{fig:difficulty}
\end{figure}

\vspace{.2mm} \noindent \textbf{Sources of Classification Difficulty.} Many rare dialogue acts have low F1 scores. Examples include \textit{Tag Question}, \textit{Commitment}, and \textit{NonSpeech}. This pattern is consistent with the limited number of examples available for these labels.

Frequency is not the only source of difficulty. Several common acts also have low F1 scores, including \textit{Defending/Explanation}, \textit{Open-ended Question}, \textit{Accept}, and \textit{Understanding Check}. These labels have more examples but remain close in meaning to other labels in the same family.

The two analyses point to different problems. Rare labels may need more training examples. Common but ambiguous labels may instead need clearer decision rules, longer dialogue context, or models that better represent the relation between an utterance and the surrounding turns. Adding more examples alone may not solve errors caused by overlapping label definitions.

\vspace{.2mm}
\noindent \textbf{Discussion.}
Our results show that LLMs do not reliably understand dialogue at the level required for fine-grained clinical analysis. Models often recognize whether an utterance is a statement, a question, or a form of feedback, but they struggle to distinguish among closely related functions within these groups. The strongest Macro-F1 is only .207, despite an Accuracy of .48. This shows that models rely on common dialogue patterns and do not consistently capture the specific communicative intent of each turn.

Better training helps, but it does not solve this problem. Instruction tuning improves Accuracy and patient-response matching, while reasoning-aware fine-tuning improves the LLaMA classification baseline. However, neither approach produces strong performance across all 56 dialogue acts. Reasoning supervision also does not consistently outperform standard instruction tuning. Future work should therefore focus on training methods that model speaker roles, longer conversational context, turn relationships, and the boundaries between similar dialogue acts. Additional in-domain training and clinically informed label definitions may also be needed.

These limitations are especially important in cognitive assessments, where test administration occurs through interaction rather than through fixed questions alone. Prior work shows that differences in how clinicians introduce, reformulate, and administer cognitive test items can affect patient understanding and responses~\cite{jones2019variation}. Interactional patterns during memory-clinic assessments, including response timing, handling of compound questions, and displays of working memory, may also provide clinically relevant information~\cite{jones2016conversational}. Clarification and conversational repair are central to identifying and resolving misunderstandings in clinician--patient communication~\cite{mccabe2018miscommunication}. More broadly, clinical communication can affect care through shared understanding, decision quality, therapeutic relationships, and patient participation~\cite{street2009does}. Models should therefore not be used to evaluate clinicians, infer cognitive status, or support clinical decisions until they can more reliably distinguish these interactional functions. Future studies should compare model errors with annotator disagreement, evaluate broader dialogue-act families, and test whether interaction patterns are associated with assessment quality, patient understanding, or clinical outcomes.

The patient-generation results should be interpreted only as reference matching. BLEU, ROUGE, and BERTScore do not show that a generated response is clinically plausible, safe, or representative of a person with a particular cognitive condition. However, they are a good initial test bed for understanding how well LLMs can simulate the utterances of cognitively impaired patients. Future work should continue to improve these. But, more importantly, clinical experts must evaluate these properties before generated responses are used in simulated-patient education. Future work should also test for inaccurate or stereotyped representations of people with cognitive impairment.

\vspace{.2mm} \noindent \textbf{Dataset Release.} We will release the annotation guidelines, label definitions, patient-level task splits, prompts, training code, evaluation code, and model outputs. Because the transcripts contain sensitive clinical conversations, the de-identified transcripts will not be posted publicly. They will be provided to researchers who submit documentation of IRB approval or a formal determination that IRB review is not required, subject to the applicable data-use agreement. We will release a standard request form to make this process clear and consistent. Our goal is to provide the data to all qualified researchers who complete these requirements.

\section{Conclusion}

We evaluated prompting, instruction tuning, and reasoning-aware fine-tuning for dialogue-act classification and patient utterance generation in cognitive assessments. Adaptation improved some results, but models still recognized broad dialogue categories more reliably than fine-grained communicative functions. Future work should study longer context, clinically informed labels, in-domain supervision, and expert evaluation before clinical use.

\bibliographystyle{aaai}
\bibliography{custom}

\end{document}